# USING LARGE LANGUAGE MODELS TO TRANSLATE MACHINE RESULTS TO HUMAN RESULTS

*Trishna Niraula, Jonathan Stubblefield*

Arkansas State University, Jonesboro, AR, USA

## ABSTRACT

Artificial intelligence (AI) has transformed medical imaging, with computer vision (CV) systems achieving state-of-the-art performance in classification and detection tasks. However, these systems typically output structured predictions, leaving radiologists responsible for translating results into full narrative reports. Recent advances in large language models (LLMs), such as GPT-4, offer new opportunities to bridge this gap by generating diagnostic narratives from structured findings.

This study introduces a pipeline that integrates YOLOv5 and YOLOv8 for anomaly detection in chest X-ray images with a large language model (LLM) to generate natural-language radiology reports. The YOLO models produce bounding-box predictions and class labels, which are then passed to the LLM to generate descriptive findings and clinical summaries.

YOLOv5 and YOLOv8 are compared in terms of detection accuracy, inference latency, and the quality of generated text, as measured by cosine similarity to ground-truth reports. Results show strong semantic similarity (0.88 ± 0.03) between AI and human reports, while human evaluation reveals GPT-4 excels in clarity (4.88/5) but exhibits lower scores for natural writing flow (2.81/5), indicating that current systems achieve clinical accuracy but remain stylistically distinguishable from radiologist-authored text.

***Index Terms***— Medical imaging, LLM, Automated report generation, Object detection, Human evaluation

## 1. INTRODUCTION

Deep learning has transformed medical imaging by providing automated tools for disease screening and diagnosis, with computer vision (CV) systems now achieving expert-level performance in detecting abnormalities from radiographic images. However, a critical gap remains between machine-generated outputs and the narrative reasoning clinicians use to communicate diagnostic findings. While CV models [3,13,14] excel at producing structured predictions such as bounding boxes, class labels, and confidence scores, these outputs cannot be directly integrated into clinical workflows, where radiologists communicate through standardized prose reports containing FINDINGS and IMPRESSION sections. This translation gap creates a bottleneck: highly accurate algorithmic detections remain inaccessible to clinicians who must manually convert structured predictions into interpretable diagnostic narratives.

The manual synthesis of computer vision outputs into radiology reports is time-consuming [22] and introduces potential inconsistencies in documentation. As medical imaging volumes continue to grow and radiologist shortages persist globally [21], automated report generation could substantially improve workflow efficiency while maintaining diagnostic quality. However, such systems must produce not merely factually accurate content but also clinically appropriate, readable, and trustworthy text that aligns with established medical communication standards. Previous approaches to automated report generation [19,23] have primarily relied on end-to-end neural architectures trained to map images directly to text, often requiring extensive paired image-report datasets and struggling to provide interpretable intermediate representations.

Recent advances in large language models (LLMs) [7,17,24], particularly GPT-4, offer new opportunities to bridge the gap between structured computer vision outputs and natural-language diagnostic narratives. These models demonstrate remarkable capabilities in generating coherent, contextually appropriate text across diverse domains, including specialized technical fields. Moreover, the emergence of vision-enabled LLMs [24,25] introduces the possibility of multimodal input, allowing models to process both visual information and textual prompts simultaneously. This capability suggests a promising architecture: using established object detection models to identify and localize pathologies, then leveraging LLMs to translate these structured findings into clinically meaningful narrative reports.

Building on our preliminary work [26], this study introduces and evaluates a two-phase pipeline that integrates YOLO-based object detection with large language models for automated radiology report generation from chest X-rays. The first phase employs YOLOv5 for anomaly detection with GPT-3.5 for text-based report generation, evaluated through quantitative semantic similarity analysis. The second phase advances this approach by combining YOLOv8 detection with GPT-4's vision capabilities, enabling direct processing of annotated images and evaluated through IRB-approved human assessment. This dual-phase design allows systematic comparison of text-based versus vision-enabled approaches while examining both semantic accuracy and clinical interpretability.

The key contributions of this work are threefold. First, we demonstrate a practical pipeline architecture that successfully translates computer vision detections into clinically interpretable radiology reports, achieving strong semantic alignment with ground-truth clinician-authored text (cosine similarity: 0.88 ± 0.03). Second, we provide a comprehensive evaluation framework combining quantitative semantic metrics with rigorous human assessment, revealing that while current LLMs excel in clarity and content accuracy, stylistic differences from human radiological writing remain apparent. Third, we offer empirical insights into the evolution from text-based to vision-enabled multimodal approaches, documenting both the capabilities and current limitations of LLM-based medical report generation.

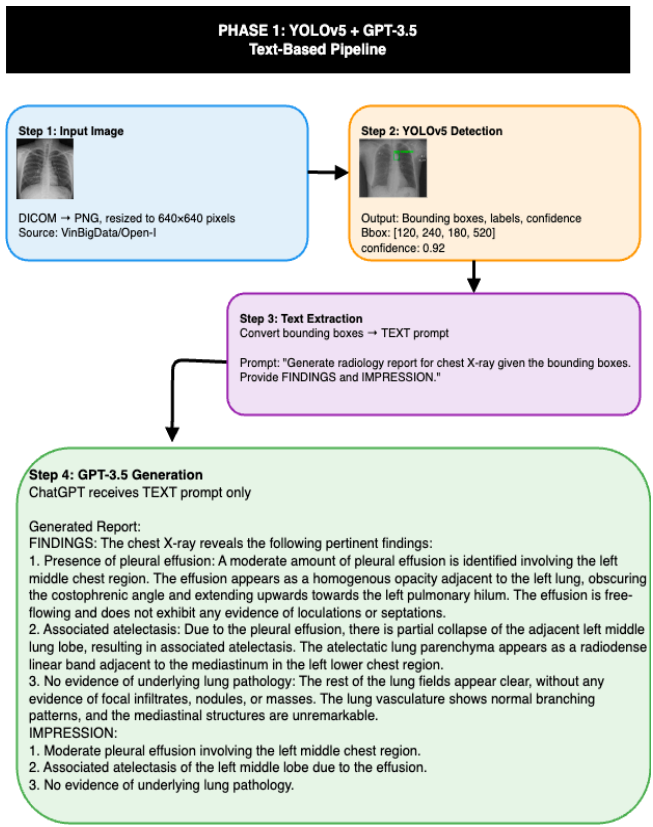

Figure 1: Overview of YOLOv5 and LLMs

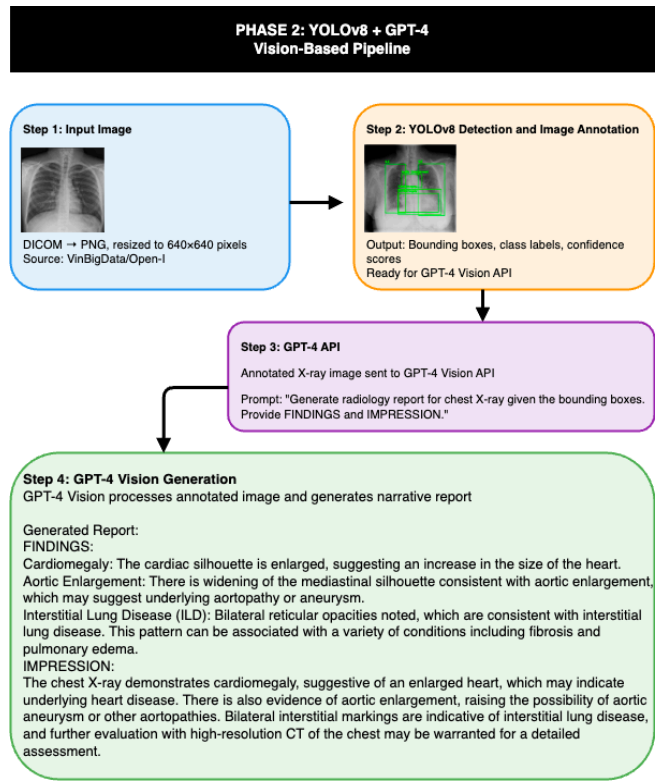

Figure 2: Overview of YOLOv5 and LLM

The remainder of this paper is organized as follows. Section 2 reviews related work in automated medical image interpretation and report generation. Section 3 describes the datasets, model architectures, and evaluation methodologies for both phases. Section 4 presents quantitative and qualitative results from cosine similarity analysis and human evaluation. Section 5 discusses implications for clinical deployment, methodological insights, and limitations. Section 6 concludes with a summary of findings and directions for future work.

## 2. Related Works

### 2.1 Automated chest X-ray Interpretation

Recent advances in deep learning have enabled automated chest X-ray analysis at expert-level performance. CheXNet demonstrated that convolutional neural networks could exceed radiologist performance on pneumonia detection, while CheXpert introduced a large-scale dataset with uncertainty labels and established benchmark performance across 14 pathology classes. The MIMIC-CXR and Open-I dataset [5,8] paired chest radiographs with free-text radiology reports, enabling research on joint vision-language tasks.

Object detection models such as YOLO [3] and Faster R-CNN have been successfully adapted for medical imaging, providing precise localization of abnormalities through bounding box predictions. YOLOv5 and YOLOv8 [1,2] represent recent advances in this family, and offer improved accuracy and efficiency. While these models achieve high accuracy in identifying and localizing pathologies, they produce structured outputs that require manual interpretation and translation into clinical narratives.

**2.2 Automated Radiology Report Generation**

The task of generating natural language reports from medical images has been approached through various architectures. Encoder-decoder models with attention mechanisms [19,23] have shown promise in generating coherent reports, learning to map visual features directly to textual descriptions. Memory-driven transformers [10] incorporate external knowledge to improve clinical accuracy and reduce hallucinations in generated text.

However, these end-to-end approaches face several limitations. They require large paired image-report datasets for training, often struggle with rare pathologies, and provide limited interpretability in terms of which visual features drive specific textual outputs. Template-based methods provide more control but lack the flexibility to handle diverse clinical scenarios and nuanced findings.

**2.3 Large Language Models for Medical Applications**

Recent advances in Large language models (LLMs) [7,17,24], particularly GPT-4 have given strong performance across diverse medical reasoning tasks, while specialized models like BioGPT [16] have been fine-tuned on medical corpora to improve domain-specific accuracy. Recent surveys [9] have demonstrated the growing application of LLMs across various medical domains, from clinical decision support to medical education.

Recent work has begun exploring LLMs for radiology report generation. Some approaches use GPT models to refine or expand template-based outputs, while others utilize prompting strategies to generate reports from textual descriptions of findings. However, most prior work treats report generation as a pure text-to-text task, not fully utilizing the visual grounding that object detection models provide.

Vision-language models such as CLIP [25] and BLIP [18] have shown promise in medical imaging by learning joint representations of images and text. GPT-4's vision capabilities extend this paradigm that enables direct processing of medical images alongside textual prompts. To our knowledge, limited work has systematically explored how vision-enabled LLMs perform on medical report generation compared to text-only approaches.

**2.4 Evaluation of Medical AI Systems**

Evaluating AI-generated medical text presents unique challenges. Automated metrics like BLEU and ROUGE, commonly used in natural language generation, correlate poorly with clinical quality. Semantic similarity measures using embedding models [6] better capture content equivalence despite surface-level differences in wording. Sentence-BERT [12] and similar approaches have enabled more nuanced semantic comparison through dense vector representations. Human evaluation remains the gold standard but introduces challenges of scale, cost, and inter-annotator variability. Recent studies, including work on self-supervised learning for pathology detection [11], have employed structured evaluation frameworks where clinicians or domain experts rate generated outputs on dimensions such as clinical accuracy, completeness, and appropriateness. However, few studies have combined quantitative semantic metrics with rigorous human assessment to provide comprehensive evaluation.

**2.5 Positioning of this work**

This work differs from prior approaches in several key aspects. First, rather than end-to-end image-to-text generation, we propose a modular pipeline that explicitly separates visual analysis (YOLO [1,2,3]) from narrative generation (LLM), that provides interpretable intermediate representations. The pipeline utilizes datasets including VinBigData [4] for training and Open-I [5,8] for evaluation, that ensures diverse representation of thoracic pathologies.

Second, we systematically compare text-based (GPT-3.5) and vision-enabled (GPT-4 Vision) approaches to report generation, documenting the evolution in capabilities as LLMs incorporate multimodal understanding. Third, we employ a dual evaluation methodology combining quantitative semantic similarity (using OpenAI's embedding model [6]) with IRB-approved human assessment.

The two-phase design provides systematic comparison of architectural choices (text-only vs. vision-enabled) and evaluation methodologies (semantic similarity vs. human judgment), contributing both a practical system and methodological insights for future research in explainable medical AI [9,10,11]. As large language models continue to advance [7], this framework provides a foundation for measuring progress and identifying areas requiring continued technical development.

### 3. METHODS

**3.1 Dataset and Preprocessing**

Two publicly available chest X-ray datasets were used in this study. The VinBigData Chest X-ray Abnormalities Detection dataset was our training set for the computer vision model. It includes over 15,000 images, each labeled with one or more of 14 thoracic abnormalities, which makes it well-suited for supervised object detection training.

The VinBigData Chest X-ray dataset was used as the primary training dataset due to its detailed bounding-box annotations for 14 thoracic abnormalities.(DICOM → PNG) Images were resized to 640 × 640 pixels, normalized, and split into patient-wise training, validation, and test sets to avoid overlap.

For inference and evaluation, we used the Open-I chest X-ray dataset (Indiana University Collection) from the U.S. National Library of Medicine. This dataset contains chest radiographs along with ground-truth radiology reports written by clinicians.

### 3.2 Evaluation

The Ultralytics open-source framework was used to implement both YOLOv5 and YOLOv8 models for automated detection of thoracic abnormalities in chest X-rays. Each model used a YAML configuration file that specified the dataset structure, training and validation splits, and class mappings. The YAML file listed 14 thoracic abnormality categories, such as cardiomegaly, pleural effusion, lung opacity, atelectasis, and aortic enlargement. Using the same YAML definitions for both models kept class indexing consistent and made detection outputs compatible with the downstream large-language-model pipeline.

For the YOLOv5 phase, we started with pretrained Ultralytics weights and fine-tuned the model on the VinBigData Chest X-ray Abnormalities Detection dataset. Training and validation took place on the Pete Supercomputer at Oklahoma State University. We set the image resolution to 640 × 640 pixels, used a batch size of 16, a learning rate of 0.01, and trained for 300 epochs. The experiments ran in a controlled HPC environment with Python 3.10, PyTorch 2.1, OpenCV, and Ultralytics YOLOv5.

During training, performance metrics such as precision, recall, and mean Average Precision (mAP@0.5 and mAP@0.5:0.95) were logged at each epoch and visualized these metrics using YOLO's built-in monitoring tools, including precision-recall curves, F1 curves, and normalized confusion matrices. Figure 3 shows the detection accuracy for each class and highlights the model's strong sensitivity to pleural effusion and cardiomegaly.

For inference, the trained YOLOv5 model processed chest X-ray images from the Open-I dataset, generating bounding-box detections and labeled overlays. Each detection included the predicted abnormality label, confidence score, and coordinates, which were exported with the related radiology report metadata. These structured outputs were then used for GPT-based report generation and cosine-similarity analysis comparing GPT-generated and clinician-written reports.

In the YOLOv8 phase, we followed an identical training procedure. Starting with pretrained weights, we fine-tuned the model on the VinBigData dataset using the same YAML configuration for 300 epochs with 640 × 640 pixel resolution. After training, the model was validated and exported to ONNX format for optimized inference. The trained YOLOv8 model generated detections that were sent directly to GPT-4 Vision through the OpenAI API to produce diagnostic narratives. The GPT-4-generated reports were then evaluated in an IRB-approved human-subject survey for clarity, trustworthiness, and natural flow, as described in Section 3.3.

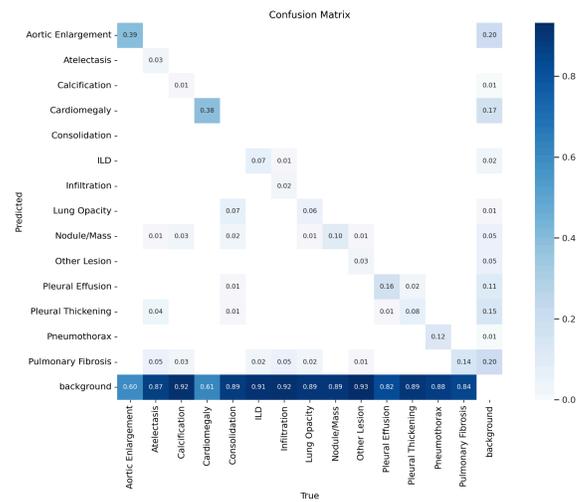

**Fig. 3. Confusion Matrix**

### 3.3 Human Evaluation and Ethical Considerations

Alongside the quantitative similarity analysis in the YOLOv5 phase, we ran a human-subject evaluation for the YOLOv8 + GPT-4 pipeline. This helped us judge how understandable and clinically useful the automatically generated radiology reports were.

This study was reviewed and approved by the Institutional Review Board (IRB) at Arkansas State University, and participation was conducted under institutional ethical guidelines.

The survey was distributed through the Department of Computer Science and the College of Engineering and Computer Science, allowing voluntary participation from individuals with backgrounds in healthcare, computer science, and engineering.

Each participant accessed the study through a secure departmental web link and reviewed a randomized set of chest-X-ray and report pairs using a custom online survey interface. Images and their corresponding GPT-generated reports were displayed in an alternating zigzag pattern to minimize positional and visual bias.

Each session presented at least two unique image–report samples, rotating automatically every minute to ensure exposure to varied cases.

For each image and report pair, participants rated the GPT-generated reports using three main criteria:

**Accuracy of findings:** how well the reported abnormalities matched what was visible in the image;

**Readability and coherence:** the clarity, organization, and language quality of the report; and

**Clinical plausibility:** whether the described conditions seemed appropriate and realistic from a diagnostic perspective.

Participants rated each report on a five-point Likert scale. They could also leave open-ended comments to share their thoughts. The ratings were averaged for each criterion and the comments were grouped by theme to identify common strengths and weaknesses in the GPT-generated clinical text.

All data were anonymized prior to analysis, and no information that could identify patients or participants was collected. This human evaluation provided additional qualitative evidence that the language model can produce radiology reports that are comprehensible and are clinically useful.

### 4. EXPERIMENT AND RESULT

#### 4.1 YOLOv5

In this phase, YOLOv5 and ChatGPT (GPT-3.5) were used to compare how closely AI-generated radiology reports matched those written by clinicians, using cosine similarity as the metric. Embedding vectors for both sets of reports were generated using OpenAI's text-embedding-ada-002 model and compared with cosine similarity to capture semantic overlap. Similarity scores ranged from 0.804 to 0.934, with a mean of 0.88 ± 0.03. This indicates strong agreement between GPT-generated and ground-truth reports.

Higher similarity scores were observed for images with clear, high-contrast conditions like cardiomegaly or pleural effusion. Lower scores occurred when images showed ambiguous or overlapping abnormalities. Although there were small wording differences, the GPT-generated reports matched the correct clinical context. This shows that a language model can accurately describe features detected by the machine.

**Figure 4** illustrates the distribution of cosine similarity scores across all test samples. **Figure 5** presents an example chest radiograph with YOLOv5-detected pleural effusion marked by a green bounding box and label, the corresponding human-written report **(Figure 5a)**, and the GPT-generated report with FINDINGS and IMPRESSION sections **(Figure 5b)**.

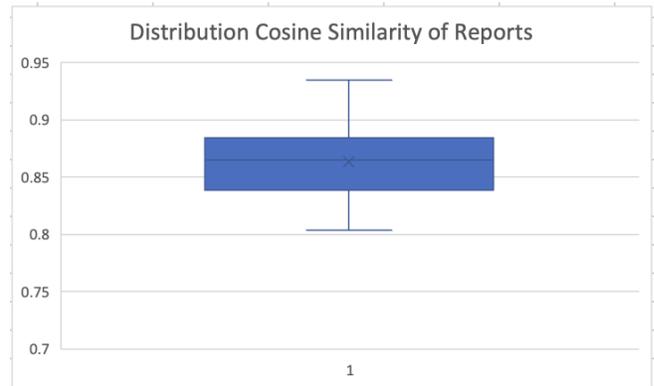

**Fig. 4.** Distribution of cosine similarity scores between GPT-generated report and human-written radiology reports. Box represents interquartile range (Q1≈0.84, Q3≈0.88), center line shows median (≈0.87), whiskers extend to minimum (≈0.81) and maximum (≈0.94). Mean: 0.88±0.03.

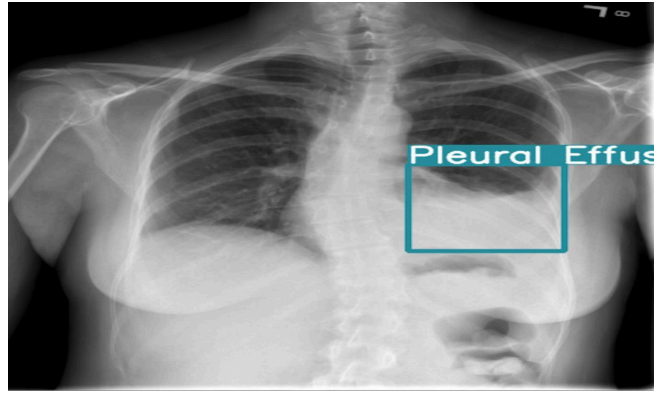

**Fig. 5.** Annotated chest X-ray

FINDINGS:
Heart size and mediastinal contour are normal. Pulmonary vascularity is normal. The right lung is clear. There is a recurrence moderate-sized left pleural effusion. No pneumothorax. Limited right base stringy density compatible with atelectasis. Dextroscoliosis of the thoracic spine.

IMPRESSION:
Recurrent moderate sized left pleural effusion.

**Fig. 5a.** Human-written radiology report

FINDINGS:
The radiographic evaluation reveals the presence of a pleural effusion in the right middle lung field. Image analysis demonstrates a collection of fluid in the pleural space between the lung and the chest wall, indicative of pleural effusion. The effusion appears well-demarcated on both the frontal and lateral projections, obscuring the normal lung markings in the affected region.

IMPRESSION:
1. Pleural effusion in the right middle lung field, as described above.
2. Absence of associated lung parenchymal collapse or significant mediastinal displacement.

**Fig. 5b.** GPT generated radiology report

**Note:** Figure 5a shows a case where the human radiologist found a left pleural effusion, but GPT described it as right-sided. This difference comes from how the image orientation is handled, since medical images can be shown in different coordinate systems such as patient-oriented or viewer-oriented.

The anatomical location error in Figure 5b ("right middle lung field") shows a fundamental limitation of the text-based pipeline. YOLOv5 detection bounding boxes are specified in pixel coordinates (x, y, width, height), which must be translated to anatomical terminology for report generation. The pipeline converted bounding box vertical position to spatial descriptors ("upper," "middle," "lower") based on Y-coordinate ranges without pathology-specific anatomical reasoning. When YOLO detected a pleural effusion with a bounding box whose vertical center fell in the middle third of the image (in pixel space), the system described it as "middle lung field" despite the anatomical fact that pleural effusions accumulate at lung bases due to gravity.

## 4.2 YOLOv8

During the YOLOv8 and GPT-4 phase, evaluation shifted from quantitative similarity metrics to human judgment through an IRB-approved survey. This blind comparison study assessed whether human evaluators could distinguish between AI-generated and clinician-written radiology reports.

*4.2.1 Survey Design and Methodology*

The evaluation used a randomized system where each participant reviewed chest X-ray images paired with radiology reports. Participants were informed that reports could be either AI-generated or human-written, but were not told which was which. The survey displayed two randomly selected image-report pairs per session, with content automatically shuffled every minute to provide a diverse set of cases and minimize position bias.

A total of 47 participants from the Department of Nursing and College of Engineering and Computer Science completed the survey, generating 94 individual report evaluations (47 participants × 2 reports each). Of these evaluations, 58 assessed GPT-generated reports and 36 assessed human-written reports. Participant backgrounds included healthcare, computer science, and engineering disciplines.

For each report, participants responded to the following evaluation criteria:

1.) Clarity (Q1): "The report's impressions and findings were presented clearly and understandably"
2.) AI Detection (Q2): "I believe the report I read was written by an AI"
3.) Trustworthiness (Q3): "I felt confident in the accuracy and trustworthiness of the information presented"
4.) Natural Flow (Q5): "The flow of the report resembled that of a human writer, with natural transitions between ideas"

Questions 1, 3, and 5 used a 5-point Likert scale (Strongly Disagree to Strongly Agree), while Question 2 was binary (Yes/No). Participants also provided qualitative feedback explaining their reasoning.

*4.2.2 Quantitative Results*

**Table 1** presents the comprehensive evaluation results comparing GPT-4-generated and human-written radiology reports across all measured criteria.

**Table 1: Human Evaluation Results - GPT-4 vs Human Reports**

| Criterion | Report Type | Mean Score | Agreement | n |
|---|---|---|---|---|
| **Clarity** | GPT-4 | 4.88/5.0 | 87.9% | 58 |
| | Human | 4.97/5.0 | 97.2% | 36 |
| **Trustworthiness** | GPT-4 | 3.36/5.0 | 65.5% | 58 |
| | Human | 4.78/5.0 | 88.9% | 36 |
| **Natural Flow** | GPT-4 | 2.81/5.0 | 37.9% | 58 |
| | Human | 4.94/5.0 | 97.2% | 36 |
| **AI Detection Accuracy** | GPT-4 | — | 70.7% | 58 |
| | Human | — | 88.9% | 36 |

*Note: Agreement represents the percentage of responses rating "Agree" or "Strongly Agree" for Likert-scale questions, and correct identification percentage for AI detection. Mean scores calculated as: Strongly Disagree=1, Disagree=2, Neutral=3, Agree=4, Strongly Agree=5.*

**Table 2** provides the detailed response distribution for GPT-4-generated reports across the three Likert-scale evaluation criteria.

Table 2: Response Distribution for GPT-4 Reports (n=58)

| Criterion | Strongly Agree | Agree | Neutral | Disagree | Strongly Disagree |
|---|---|---|---|---|---|
| Clarity | 0 (0%) | 51 (87.9%) | 7 (12.1%) | 0 (0%) | 0 (0%) |
| Trustworthiness | 0 (0%) | 38 (65.5%) | 3 (5.2%) | 9 (15.5%) | 8 (13.8%) |
| Natural Flow | 0 (0%) | 22 (37.9%) | 5 (8.6%) | 31 (53.4%) | 0 (0%) |

The results indicate a nuanced performance profile for GPT-4-generated reports. The model demonstrated high clarity (mean: 4.88/5.0, 87.9% agreement), closely matching human-written reports (4.97/5.0, 97.2% agreement), performance was more variable in trustworthiness (3.36/5.0, 65.5% agreement) and notably weaker in natural flow (2.81/5.0, 37.9% agreement). The majority of participants (53.4%) disagreed that GPT-4 reports exhibited human-like writing flow, indicating that although the content was clear and accurate, the writing style remained recognizably algorithmic.

In terms of detectability, 70.7% of participants correctly identified GPT-4 reports as AI-generated, whereas 88.9% correctly identified human reports as human-written. This suggests that while GPT-4 reports are not indistinguishable from human writing, nearly 30% of evaluators could not differentiate them from clinician-authored content.

*4.2.3 Qualitative Findings*

Participant feedback highlighted what sets GPT-4-generated reports apart. The most frequent sign of AI authorship was too much focus on detail. These reports often gave full anatomical descriptions, even when findings were normal, while human radiologists usually keep documentation brief. Participants commonly noted:

- "The report was overly detailed for normal anatomy"
- "Too structured compared to typical radiology reports"
- "Clinicians usually focus only on abnormalities, but this report described everything."
- "The language sounded clinical, but was not as brief as reports written by humans."

Some participants also pointed out positive aspects, calling GPT-4 reports "well-organized," "easy to follow," and "clinically appropriate." Several said the clear structure and thorough coverage made the reports more useful for education, even though they were less brief than typical clinical reports.

*4.2.4 Interpretation*

These results show that GPT-4 can generate radiology reports that are clear and mostly reliable, although they do not match the style of human writing. The trade-off between comprehensiveness and conciseness represents a design choice rather than a fundamental limitation, GPT-4 could potentially be fine-tuned to adopt more concise clinical language if desired. The current level of detail may actually help in educational and training settings where detailed anatomical descriptions are useful.

The use of blind comparison in this study makes the results more reliable because participants judged the reports without knowing where they came from, which removed confirmation bias. The high level of agreement on clarity for both AI and human reports (87.9% vs. 97.2%) shows that GPT-4 can communicate diagnostic information as well as humans, even though experts can still notice differences in style.

**4.3 Comparative Findings**

The two-phase experimental design shows that both quantitative and qualitative methods have unique strengths when evaluating AI-generated radiology reports. YOLOv5 with GPT-3.5 achieved high semantic similarity to ground-truth reports (mean cosine similarity: 0.88 ± 0.03). However, human evaluation of YOLOv8 with GPT-4 revealed subtle performance details that semantic metrics alone do not show.

**Table 3** summarizes the key findings from both evaluation phases.

Table 3: Comparison of Evaluation Phases

| Aspect | Phase 1: YOLOv5 + GPT-3.5 | Phase 2: YOLOv8 + GPT-4 |
|---|---|---|
| Evaluation Method | Quantitative: Cosine similarity to ground-truth reports | Qualitative: IRB-approved human evaluation with blind comparison |
| **Primary Strength** | Demonstrates semantic equivalence (0.88 mean similarity) | Excellent clarity (4.88/5) and clinical appropriateness |
| **Primary Limitation** | Cannot assess readability, writing style, or clinical plausibility | Lower trustworthiness (3.36/5) and unnatural flow (2.81/5) |
| **Key Insight** | LLMs can translate CV outputs into clinically meaningful text | GPT-4 is clear but stylistically distinguishable from human writing |

The transition from YOLOv5 combined with GPT-3.5 to YOLOv8 combined with GPT-4 represents advancements in both computer vision and language modeling. However, human evaluation indicates that improved technology does not necessarily result in outputs perceived as human-generated. While GPT-4 demonstrates clarity in presenting information (87.9% agreement on clarity), its writing style is frequently identified as artificial intelligence by human evaluators (70.7% correctly identified as AI).

A notable finding is that GPT-4 exhibits variable performance across evaluation criteria. Although it approaches human-level clarity in reports (4.88/5 compared to 4.97/5), it falls short in trustworthiness (3.36/5 versus 4.78/5) and natural flow (2.81/5 versus 4.94/5). While current language models effectively convey diagnostic information, they do not yet replicate the concise and focused writing style characteristic of experienced radiologists.

The two evaluation phases work well together. Cosine similarity measures how much the content matches, but it does not capture style, readability, or clinical naturalness. Human evaluation gives detailed feedback on these aspects, though it takes more time and can be subjective. Using both methods, first quantitative metrics, then careful human review, gives a fuller picture of system performance than either method alone.

In summary, the results indicate that the YOLO combined with a large language model (LLM) pipeline effectively bridges machine detections and human-readable reports. The system generates reports that are accurate and clear for clinical application, although the writing style remains distinct from that of human authors. This development represents progress in explainable medical artificial intelligence, while achieving human-level naturalness remains an objective for future research.

## 5. DISCUSSION

This study shows both the potential and the limits of using large language models to turn structured computer vision outputs into narrative radiology reports. We combined object detection models like YOLOv5 and YOLOv8 with generative language models such as GPT-3.5 and GPT-4 to build a pipeline that produces clinically useful text from machine-generated detections. Our two-phase evaluation, which included quantitative semantic similarity and then human assessment, highlights both the strengths and current challenges of this method.

### 5.1 Semantic Coherence vs. Stylistic Naturalness

In the YOLOv5 + GPT-3.5 phase, the model showed strong semantic alignment with ground-truth reports, with a mean cosine similarity of $0.88 \pm 0.03$. These findings indicate that large language models, when given structured visual findings, can produce text that captures the main diagnostic content found in clinician-authored reports. The overlap is highest for clear findings like cardiomegaly and pleural effusion, but it drops for more complex or subtle abnormalities, similar to how radiologists often disagree in challenging cases.

However, evaluation of YOLOv8 combined with GPT-4 revealed that high semantic similarity did not equate to reports being indistinguishable from those written by humans. Participants rated GPT-4-generated reports highly for clarity (4.88/5, 87.9% agreement), but assigned lower scores for trustworthiness (3.36/5, 65.5% agreement) and natural flow (2.81/5, 37.9% agreement). This discrepancy indicates that, although large language models can accurately convey diagnostic information, their writing style remains noticeably artificial.

Participants mainly noticed that the reports were too comprehensive. Human radiologists usually use brief language, noting only abnormal findings and leaving out normal anatomy. In contrast, GPT-4 describes every structure it sees, making the reports thorough but wordy. This happens because large language models are trained to give complete, detailed answers, not to follow the concise style used in clinical radiology.

### 5.2 Implications for Clinical Deployment

The fact that GPT-4 is clear but not always concise has real effects on how it can be used in clinics. In settings like radiology training or patient education, the detailed reports from GPT-4 can be helpful. These thorough anatomical descriptions can teach trainees how to interpret images step by step and give patients clear explanations of their results.

On the other hand, in busy clinical settings where radiologists need to work quickly, long AI-generated reports can slow things down. Radiologists who are used to short, direct reports may find these texts too wordy, which could hurt productivity. To use AI successfully in these settings, the output style will need to be adjusted to match the usual way radiologists communicate.

The average trustworthiness rating of 3.36 out of 5 is important to note. This score is more about how the reports sound than about whether they are accurate. For AI to be accepted in clinics, people need to trust its output. Even if the information is correct, clinicians may not use it if it does not seem reliable. Fixing this will need both technical upgrades and clear explanations of what the system can and cannot do.

### 5.3 Evaluation Methodology Insights

This study's methodological contribution lies in demonstrating the necessity of multi-modal evaluation for medical AI systems. Cosine similarity provided an efficient, scalable method for assessing semantic content but proved insufficient for evaluating clinical utility. The human evaluation phase highlighted performance dimensions readability, trustworthiness, stylistic naturalness that are critical for clinical acceptance but invisible to embedding-based metrics.

The blind comparison design, in which participants evaluated both AI and human reports without knowing the source, strengthens the validity of our findings by eliminating confirmation bias. The fact that 29.3% of participants could not distinguish GPT-4 from human reports shows the system has achieved meaningful capability, even if it is not perfectly indistinguishable.

Future medical AI evaluations should consider adopting similar hybrid approaches, combining quantitative performance metrics with structured human assessment. The IRB-approved survey framework used in this study provides a template for ethically conducting such evaluations while protecting participant and patient privacy.

### 5.4 Limitations

There are several reasons why these findings may not apply broadly. First, the study only looked at chest X-ray interpretation, which is a narrow area of imaging. Results could be quite different for more complex scans like CT or MRI, or for other parts of the body. Second, the human evaluation group was small, with 47 participants and 94 evaluations, and most were from technical backgrounds rather than clinical ones. If board-certified radiologists had evaluated the results, their ratings of trustworthiness and plausibility might have been different.

Third, the GPT-4 system was not specifically fine-tuned for radiology reporting. Specialized medical language models or domain-adapted prompting strategies might substantially improve performance, particularly on the natural flow and conciseness dimensions where GPT-4 currently underperforms. Fourth, we did not systematically evaluate factual accuracy against ground-truth diagnoses, instead relying on embedding similarity and human perception. A formal accuracy assessment comparing AI detections and generated text against gold-standard radiological interpretations would strengthen clinical validity claims.

Finally, this study evaluated report generation in isolation, without considering the broader clinical workflow. Real-world deployment would require integration with electronic health records, consideration of medicolegal implications, and mechanisms for radiologist review and correction of AI outputs.

### 5.5 Future Directions

This work points to several promising directions. Fine-tuning large language models on carefully selected radiology reports could help match the output style to clinical standards while keeping information accurate. Creating prompt strategies that ask for concise responses, such as "Report only abnormal findings," may help reduce unnecessary details without retraining the model. Trying hybrid methods that mix template-based structured reporting with LLM-generated narrative sections could offer both standardization and flexibility.

Trustworthiness deserves focused study. Learning which report features build clinician confidence, such as specific terms, clear structure, or careful wording for uncertain findings, could guide improvements. Adding ways for the LLM to show how confident it is in its statements may help make reports more trustworthy and safer for clinical use.

Finally, testing these systems in real clinical trials, where radiologists use AI-generated reports in their daily work, would give strong evidence of their practical value. These studies should look at not just the quality of the reports, but also results like diagnostic accuracy, reporting speed, and the overall quality of patient care.

In summary, combining computer vision and large language models can turn machine-level image analysis into clear, readable diagnostic reports. Current systems create accurate reports, but matching the concise and natural style of expert radiologists is still a challenge. The evaluation framework used here offers a solid way to measure progress and highlights where more technical work is needed.

## 6. CONCLUSION

This study demonstrates that integrating computer vision and large language models can successfully translate machine-detected abnormalities from chest X-rays into clinically interpretable radiology reports. The proposed YOLO with LLM pipeline achieves strong semantic alignment with ground-truth reports (0.88 cosine similarity) and generates text that human evaluators rate highly for clarity (4.88/5), validating the core technical approach.

The comprehensive two-phase evaluation framework, combining quantitative semantic similarity with IRB-approved human assessment, shows both capabilities and limitations. While GPT-4 effectively communicates diagnostic information with excellent clarity, current outputs remain stylistically distinguishable from expert radiologist writing, receiving moderate ratings for trustworthiness (3.36/5) and natural flow (2.81/5). The primary differentiator is verbosity: LLMs provide comprehensive anatomical descriptions whereas human radiologists employ concise, abnormality-focused language.

These findings have practical implications for deployment. The detailed outputs of large language models may help in educational settings, such as radiology training and patient communication. However, clinical workflows that need efficiency will require domain-specific fine-tuning to match the brief style typical of radiology. Notably, 29.3% of evaluators could not tell GPT-4 reports apart from human ones, even with stylistic differences. This shows real progress toward generating natural-sounding medical text.

This work makes both technical and methodological contributions. On the technical side, it sets up a working system for automated radiology report generation that creates accurate and clear narratives from computer vision results. Methodologically, it shows that a hybrid evaluation, which combines embedding-based semantic metrics with structured human review, is needed to fully assess medical AI systems. This is because numbers alone cannot measure readability, natural style, or clinical plausibility.

Although matching the concise and natural style of experienced radiologists is still a challenge, this framework offers a way to measure progress and find areas that need more work. As mentioned in Section 5.5, future research should focus on fine-tuning for specific domains, making prompts more concise, measuring uncertainty, and running clinical trials to test real-world usefulness.

This approach, which uses computer vision to extract structured findings and language models to generate narrative explanations, could be used in other medical imaging fields as well. It offers a general framework for explainable medical AI. As large language models improve, AI-assisted report generation could help clinical workflows, support medical education, and improve healthcare in settings with limited resources.

## 7. ACKNOWLEDGMENTS

This research was supported by the Data Analytics that are Robust and Trusted (DART) Program, funded by the National Science Foundation (NSF EPSCoR, Award #OIA-1946391). We thank Oklahoma State University for access to the Pete Supercomputer, which provided high-performance computing resources for model training and analysis. We also appreciate the Arkansas State University Institutional Review Board (IRB) for approving the human-subject evaluation and the participants who volunteered their feedback.